\documentclass[10pt]{cai}

\graphicspath{{figs/}}

\begin{document}

\begin{center}

  \title{Automated Journalistic Questions: A New Method for Extracting 5W1H in French}
  \maketitle

  \thispagestyle{empty}

  \begin{tabular}{cc}
    Maxence Verhaverbeke\upstairs{\affilone}, Julie A. Gramaccia\upstairs{\affiltwo}, Richard Khoury\upstairs{\affilone,*}
   \\[0.25ex]
   {\small \upstairs{\affilone} Université Laval \upstairs{\affiltwo} Ottawa University
   } \\
  \end{tabular}
  
  \emails{
    \upstairs{*}richard.khoury@ift.ulaval.ca
    }
  \vspace*{0.2in}
\end{center}

\begin{abstract}

The 5W1H questions - who, what, when, where, why and how - are commonly used in journalism to ensure that an article describes events clearly and systematically. Answering them is a crucial prerequisites for tasks such as summarization, clustering, and news aggregation. In this paper, we design the first automated extraction pipeline to get 5W1H information from French news articles. 
To evaluate the performance of our algorithm, we also create a corpus of 250 Quebec news articles with 5W1H answers marked by four human annotators. Our results demonstrate that our pipeline performs as well in this task as the large language model GPT-4o. 
\end{abstract}

\copyrightnotice

\section{Introduction}
\label{intro}

News articles inform readers about current events by providing answers to six questions: who, when, why, what, where, and how, collectively called the 5W1H questions. These are considered the foundation of journalistic work. They are taught from the very first days in journalism schools, and are the cornerstone of the inverted pyramid principle \cite{osti_10274096}. 
Extracting this key information from news articles is a widely studied task across various disciplines. For instance, social science researchers perform frame analysis \cite{goffman1974frame} to examine how media outlets describe and report events. In artificial intelligence (AI), this task is often a text preprocessing step, and serves to enhance the performance of downstream natural language processing (NLP) tasks \cite{chakma2019deep, norambuena2020evaluating}. As a result, 5W1H extraction algorithms exist in several languages,  such as English  \cite{hamborg2019giveme5w1h} and Chinese \cite{wang2010chinese,wang2010extracting}. However, no such algorithm exists in the French language, leaving only the option of querying a black-box generative AI tool to obtain answers. 
In this context, this paper makes three key contributions:
\begin{itemize}
    \item We propose the first algorithm to extract 5W1H answers from French news articles. 
    \item We construct a corpus of 249 Québec French news articles with 5W1H answers manually annotated by four independent human annotators.
    \item We demonstrate that our extraction pipelime algorithm performs as well as a generative AI tool, while being completely transparent in its decision-making.
\end{itemize}

The remainder of this paper is organized as follows. The next section provides a review of the literature on the 5W1H extraction problem. Sections \ref{methodology} and \ref{dataset} describe our algorithm and dataset, respectively, followed by the results presented in Section \ref{results}. Finally, concluding remarks are provided in Section \ref{conclusion}.

\section{Related Work}
\label{related_work}

The 5W1H are fundamental to news article writing. They ensure the clarity of the information and its comprehension by readers \cite{MartinLagardette2005}. They also serve as the guiding thread that structures the article. The core information (who does what where and when) is usually included in the lead of the article, followed by other key details (how and why) and finally by secondary information \cite{Ross2005}. This structure is known as the inverted pyramid, which forms the backbone of most  informational news stories.

Consequently, 5W1H extraction has become a cornerstone of news article processing in NLP. It is also a preprocessing step in NLP applications as varied as text structure evaluation \cite{norambuena2020evaluating}, sentiment analysis \cite{das20125w}, and tweet understanding \cite{chakma2019deep}.

\begin{figure}[ht]
    \centering
    \includegraphics[scale=0.5]{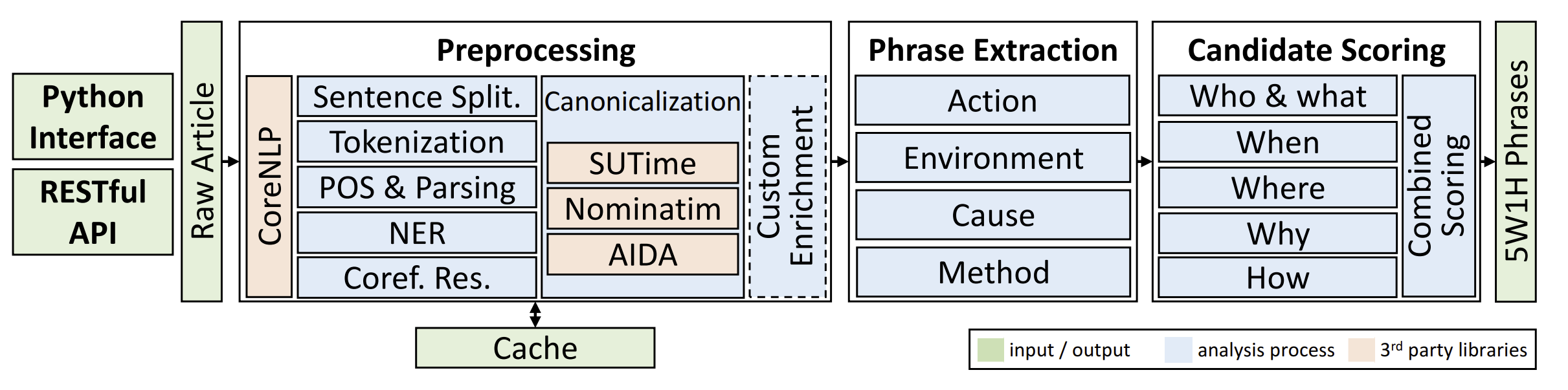}
    \caption{The Gimme5W1H pipeline (taken from \cite{hamborg2019giveme5w1h}).}
    \label{fig:g5w1hpipe}
\end{figure}

The most common approach consists of implementing a processing pipeline that gradually extracts the answers. A popular example is the Gimme5W1H pipeline \cite{hamborg2019giveme5w1h, norambuena2020evaluating}, reproduced in Figure \ref{fig:g5w1hpipe}. This pipeline begins by applying standard NLP preprocessing algorithms to clean up a news article's text, then it uses four different phrase extraction algorithms to identify candidate answers to the six questions, and finally it ranks these candidates using question-specific weighting schemes to return the best answer to each question. The authors of \cite{wang2010chinese} used a similar approach but focusing more heavily on the key sentence extraction phase, while \cite{sharma2013news} implemented simpler but separate pipelines for the different questions. 

A popular alternative to processing pipelines is building rule-based extraction systems. This approach works well thanks to the standardized and predictable structure of news articles. For example, one of the rules proposed in \cite{wang2010extracting} consists in extracting subject-verb-object triplets from the text and ordering them based on where they appear in the article in order to answer the ``who'' and ``what'' questions. The authors of \cite{lewis2019mlqa} generated POS and grammar tags for the text, then designed rules to map sentence components to 5W1H answers. Likewise, \cite{das20125w, yaman2009classification} designed rules based on standard grammar. These systems are often enhanced by machine learning: for example \cite{wang2010extracting} used an SVM to improve the answers extracted by their rules, while \cite{das20125w} used their rules to improve the answers obtained by a maximum entropy classifier. The authors of \cite{yaman2009classification} implemented three independent rule-based extractors and trained a classifier on their outputs, while \cite{chakma2019deep} trained an LSTM encoder-decoder with a middle attention layer to extract 5W1H answers. 

It is worth noting that these projects were all done in English \cite{hamborg2019giveme5w1h, chakma2019deep, norambuena2020evaluating, yaman2009classification, sharma2013news} or Chinese \cite{wang2010chinese, wang2010extracting, lewis2019mlqa}, with only one exception in Bengali \cite{das20125w}. To the best of our knowledge, no one has studied 5W1H extraction in French. While one study considered the possibility of cross-lingual 5W1H systems \cite{lewis2019mlqa}, their results were disappointing: the best cross-lingual system their tested was still $19\%$ worse than monolingual systems \cite{lewis2019mlqa}. This result is not unique to 5W1H answering, as other cross-lingual question-answering (Q\&A) systems also show drops in performances compared to their mono-lingual equivalents \cite{parton2009comparing}. 

\section{Methodology}
\label{methodology}
\subsection{Overview of the System}
\begin{figure}[ht]
    \centering
    \includegraphics[scale=0.2]{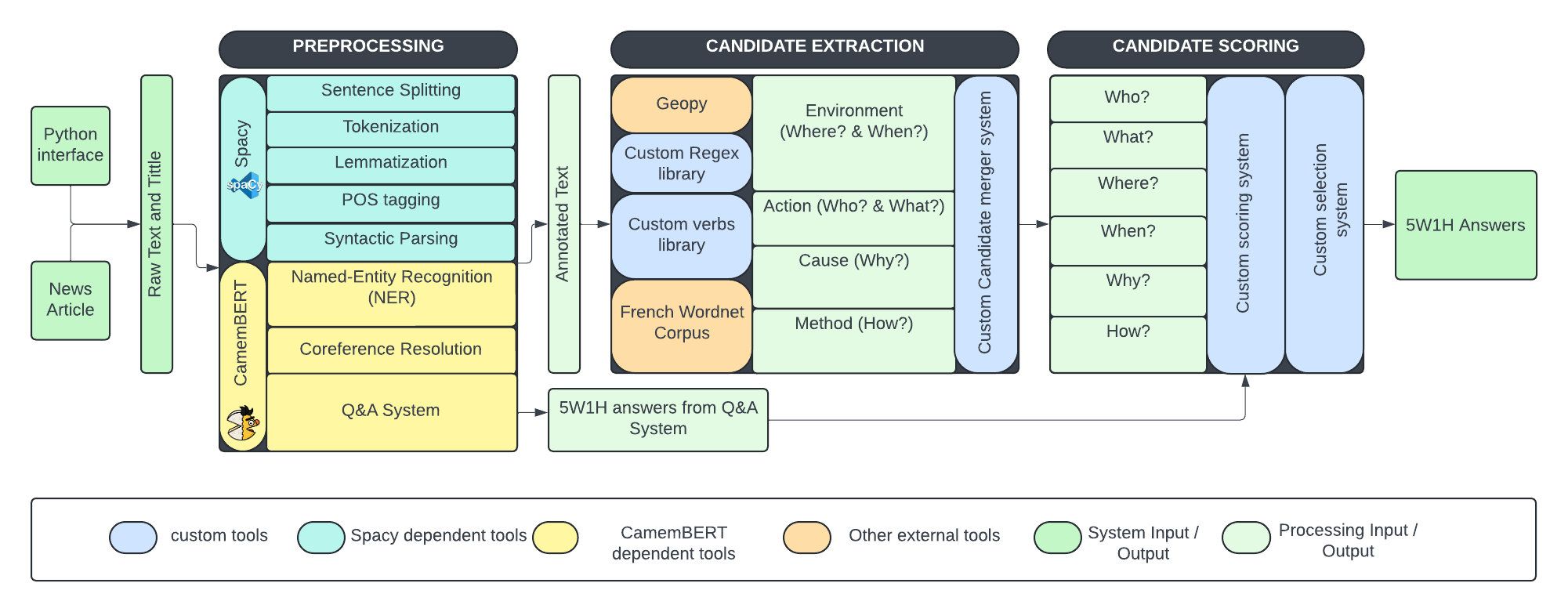}
    \caption{Our French-language 5W1H pipeline.}
    \label{fig:workflow}
\end{figure}

The French 5W1H system we propose is a modular extraction pipeline based on the one first suggested by \cite{hamborg2019giveme5w1h} for the English language. Much like that pipeline, our proposed pipeline is composed of three phases: first a preprocessing phase to clean-up and enrich the article text, next a sentence extraction phase to obtain candidate answers to the 5W1H questions, and finally a scoring phase to pick the best answers to each question. The complete pipeline is represented in Figure \ref{fig:workflow}.
The algorithm is available on our GitHub account\footnote{Removed for peer-review.}.

\subsection{Preprocessing Phase}
Our preprocessing phase uses a mix of Spacy's NLP tools and pre-trained CamemBERT Transformer models  \cite{DBLP:journals/corr/abs-1911-03894}. 
Specifically, we leverage SpaCy's ``fr\_core\_news\_lg'' model for basic NLP preprocessing operations, namely sentence splitting and tokenization, word lemmatization, part-of-speech (POS) tagging and syntactic parsing. 
Meanwhile, the CamemBERT-based models perform more sophisticated operations, namely named entity recognition (NER) and classification into five classes (person, organization, location, date, miscellaneous), and coreference resolution.

One novelty of our pipeline in this phase is the integration of a Q\&A module\footnote{\href{https://huggingface.co/AgentPublic/camembert-base-squadFR-fquad-piaf}{"etalab-ia/camembert-base-squadFR-fquad-piaf"}}, which is used to extract an initial set of candidate answers to the 5W1H questions. We provide this module a set of generic prompts and a set of prompts that allow the use of previously-discovered answers. Any answer returned with a confidence score $\geq 0.5$ is retained and incorporated into subsequent prompts, and if none are retained then the generic prompts are used. The prompts are provided in the Appendix.

\subsection{Candidate Extraction Phase}


At the candidate extraction stage, our pipeline splits into four independent modules: the action extraction module identifies candidate answers to the ``who'' and ``what'' questions; the environment module extracts candidates for ``when'' and ``where''; the cause module discovers answers to ``why''; and the method module identifies answers to ``how''.

The \textbf{action extractor} determines ``who did what'' in a news article by analyzing the output of the NER and syntactic parser from the preprocessing phase. It begins by creating a list of ``who'' candidates by grabbing the list of named entities classified as persons or organizations during preprocessing. From the syntactic parser, it also gets the list of sentences whose root is a verb phrase (VP) and adds the preceding nominal phrase (NP) subject of this root to the list of candidates. Next, the extractor checks if pairs of candidates are similar enough to be considered equivalent and keeps only one copy. 
We compare them by computing the ratio of words with more than three letters that are present in both candidates to the total number of words of more than three letters in both responses. If it exceeds a predefined threshold, the candidates are considered identical.

For the extraction of ``what'' candidates, we start with the technique laid out in \cite{hamborg2019giveme5w1h}: given a VP sentence root, all following NPs and VPs are assigned as candidate answers. 
However, we noticed this technique often fails to recognize candidate sentences when they contains quotes or when they do not follow the classic NP-VP-NP structure. We thus improve on the technique in two ways. First, we build an action verb list by taking a list of common French action verbs and adding their WordNet synonyms, and we extract all sentences that use these verbs. And second, since the title of a news article often summarizes what the article is about, any sentence similar to the title using our word ratio will be included as a candidate.


The \textbf{environment extractor} identifies the temporal and spatial context of the event. To achieve this, it first gets the list of named entities classified as locations or dates during preprocessing. However, the NER fails to recognize complex or vague temporal entities, such as \textit{d'ici un mois ou deux} (within a week or two) or \textit{dans le courant de la semaine} (sometime during the week). 
Consequently, we supplement it with a temporal entity detection system based on a regex library \footnote{\href{https://github.com/bear/parsedatetime}{bear/parsedatetime}}. This system can detect a greater variety of temporal entities, including times, dates, durations and recurring intervals. All of them are added to our list of potential candidates. 
Finally, temporal entities that are adjacent or separated only by a conjunction or punctuation are merged into a single entity. 



The \textbf{cause extractor} identifies linguistic features indicative of causal relationships using two complementary approaches. First, we provide the algorithm with a list of French causal verbs, augmented by their WordNet synonyms. When one of these verbs is found in a sentence, the complete sentence is added as a candidate. Second, the algorithm detects if a sentence contains causal markers, such as \textit{car} (because)\textit{parce que} (because), \textit{en raison de} (because of) or \textit{car} (because). When such markers are found, the sentence is also retained.

The \textbf{method extractor} identifies ``how'' candidates, representing the method by which the news event was carried out. It employs two approaches. First, we detect sentences containing verbs in the present participle (e.g. \textit{en faisant} (doing), \textit{en écrivant} (writing)), which are strong markers of explanatory phrases. This is done by getting verbs from the preprocessing POS tagger, and verifying that the preceding word is the preposition \textit{en} and the last three letters of the verb are \textit{ant}. These sentences are extracted as ``how'' candidates. The second approach finds sentences that contain prepositions, modal adverbs, or specific expressions (such as \textit{grâce à} (thanks to) or \textit{au moyen de} (using)), and adds them to the candidate list. These insure our algorithm doesn't miss any answer candidates.

\subsection{Candidate Scoring Phase}
After the previous phase has supplied the pipeline with all possible 5W1H candidate answers, this phase ranks them in order to return only the best answers to each question. 

The candidate answers of each question type are evaluated based on question-specific factors, which are detailed below. 
Each factor $f_i$ is assigned a weight $w_{iq}$ based on its importance to each question type $q$. The candidate $c$'s final score as an answer to that question is thus calculated as the weighted sum of its scores $s_{iqc}$ for each factor:
\begin{equation}
S_{cq} = \sum_{i=0}^{n-1} w_{iq}s_{iqc}
\end{equation}
The weights $w_{iq}$ are based on the values published in the original code of \cite{hamborg2019giveme5w1h} and adjusted empirically. The lack of a large amount of annotated data and the variability in responses among human annotators limited our ability to train the model to adjust the weights automatically. However, the code structure makes it easy to modify these weights and even add new factors for future improvement, when more labeled data becomes available. 

\subsubsection{``Who'' Candidates}

\begin{table}[H]
\centering
\caption{Weights and scoring factors for ``who'' candidates.}
\renewcommand{\arraystretch}{1.4} 
\setlength{\tabcolsep}{10pt} 

\begin{tabular}{>{\centering\arraybackslash}m{3cm}|m{1.5cm}| m{3cm}}
\hline
\textbf{$f_i$} & \textbf{$W_{iq}$} & \textbf{$S_{iqc}$} \\
\hline
Frequency & 0.40 & $\frac{\text{coref\_count}(c)}{\max_{c' \in C}(\text{coref\_count}(c'))}$ \\
Position & 0.25 & $1 - \frac{Position(c)}{N_\text{sentence}}$ \\
Title Presence & 0.20 & $0$ or $1$ \\
PER Type & 0.10 & $0$ or $1$\\
Q\&A Similarity & 0.05 & $0$ or $1$ \\
\hline
\end{tabular}
\label{tab:weights_scoring}
\end{table}


The scoring of ``who'' candidates is primarily based on the frequency with which it is referenced in the text: the more it appears, the more likely it is to be a central subject of the article. Using a CamemBERT corefrence resolution algorithm\footnote{\href{https://huggingface.co/Easter-Island/coref_classifier_ancor}{Easter-Island/coref\_classifier\_ancor}}, we obtain the number of coreferences of the candidate $coref\_count(c)$ and normalize as the ratio to the candidate with the maximum number of coreferences. 
Professional news articles follow a standardized structure, where the most important information is presented in the beginning of the text, then additional details and contextual information follows in the rest of the text \cite{bleyer1923newspaper}. Consequently, we dedine the candidate's normalized position as the ratio between the index position of the sentence in which it first appears $Position(c)$ and the total number of sentences in the news article $N_\text{sentence}$. 
We also checks if the candidate in present in the article's title and if the type of named entity it contains is a person (PER), both good indicators of the candidate's relevance. 
Finally, we check whether the candidate is similar to the ``who'' candidate identified by the Q\&A system in the preprocessing phase, using the same technique as in the action extractor. 

\subsubsection{``What'' Candidates}

\begin{table}[H]
\centering
\caption{Weights and scoring factors for ``what'' candidates.}
\renewcommand{\arraystretch}{1.4} 
\setlength{\tabcolsep}{10pt} 
\begin{tabular}{>{\centering\arraybackslash}m{5cm}|m{1cm} | m{3cm}}
\hline
\textbf{i} & \textbf{$W_qi$} & \textbf{$S_qi$} \\
\hline
Position & 0.50 & $1 - \frac{Position(c)}{N_\text{sentence}}$ \\
Length & 0.15 & $\frac{len(c)}{\max_{c' \in C}(len(c')}$ \\
Average Score of ``who'' Candidates & 0.15 & $\frac{\sum \text{who\_score}}{\text{who\_count}}$ \\
Action Verbs & 0.08 & $0$ or $1$ \\
NP-VP-NP Sentence & 0.07 & $0$ or $1$ \\
Q\&A Similarity & 0.05 & $0$ or $1$ \\
\hline
\end{tabular}
\label{tab:weights_scoring}
\end{table}

The scoring of ``what'' candidates predominently considers whether the candidate occurs in the first sentences of the article, where the answer to this question is typically presented \cite{goffman1974frame}. To calculate the candidate's position score, we use the same formula as for scoring ``who'' candidates. 
Additionally, we want the action described to be sufficiently detailed. Thus, to avoid capturing short introductory sentences sometimes found at the beginning of paragraphs, we take into account the length of the candidate $len(c)$ as a ratio of the longest candidate. 
The scoring of ``what'' candidates presented in \cite{hamborg2019giveme5w1h} was solely based on the results of the ``who'' candidates: the verbal group of the sentence containing the best ``who'' candidate was systematically selected as the best ``what'' candidate. However, we found that this strategy fails in the case where the article begins by introducing the ``who'' subject before presenting the actions they performed. 
We thus devised a different strategy: we compute the average score of ``who'' candidates that occur in the same sentence as the ``what'' candidate. 
Finally, we consider three binary factors. First,  we use the French WordNet corpus to identify the most common action verbs, and check whether one of these verbs appears in the candidate's sentence. 
Next, since the candidate should be an action performed by a subject, we check whether or not the candidate's sentence uses a noun phrase - verb phrase - noun phrase (NP-VP-NP) structure. 
Finally, we measure the similarity to our Q\&A module's answer to this question.


\subsubsection{``When'' Candidates}
\begin{table}[H]
\centering
\caption{Weights and scoring factors for ``when'' candidates.}
\renewcommand{\arraystretch}{1.4} 
\setlength{\tabcolsep}{10pt} 
\begin{tabular}{>{\centering\arraybackslash}m{3cm}|m{1.2cm} | m{6.5cm}}
\hline
\textbf{i} & \textbf{$W_qi$} & \textbf{$S_qi$} \\
\hline
Temporal Precision & 0.40 & time$ = 1.00$; date$ = 0.66$; set$ = 0.33$; duration$ = 0.00$ \\
Frequency & 0.30 & $\frac{\text{count}(c)}{\max_{c' \in C}(\text{count}(c'))}$ \\
Position & 0.25 & $1 - \frac{Position(c)}{N_\text{sentence}}$ \\
Q\&A Similarity & 0.05 & $0$ or $1$ \\
\hline
\end{tabular}
\label{tab:weights_scoring}
\end{table}

In addition to previously-explained factors
, the scoring of ``when'' candidates considers the precision of the identified temporal entity, and gives preference to the most precise entity.

\subsubsection{``Where'' Candidates}
\begin{table}[H]
\centering
\caption{Weights and scoring factors for ``where'' candidates.}
\renewcommand{\arraystretch}{1.4} 
\setlength{\tabcolsep}{10pt} 
\begin{tabular}{>{\centering\arraybackslash}m{3cm}|m{1.5cm} | m{5cm}}
\hline
\textbf{i} & \textbf{$W_qi$} & \textbf{$S_qi$} \\
\hline
Position & 0.32 & $1 - \frac{Position(c)}{N_\text{sentence}}$ \\
Frequency & 0.30 & $\frac{\text{count}(c)}{\max_{c' \in C}(\text{count}(c'))}$ \\
Containment & 0.30 & $\frac{\text{areas\_contained}(c)}{\max_{c' \in C}(\text{areas\_contained}(c'))}$ \\
Size & 0.03 & $1 - \min (1, \frac{\log area(c) - \log area_{\min}}{\log area_{\max} - \log area_{\min}})$ \\
Q\&A Similarity & 0.05 & $0$ or $1$ \\
\hline
\end{tabular}
\label{tab:weights_scoring}
\end{table}

Like temporal candidates, geographic candidates are evaluated based on their precision. Similarly to \cite{hamborg2019giveme5w1h}, we use Geopy\footnote{\href{https://geopy.readthedocs.io/en/stable/}{https://geopy.readthedocs.io/en/stable}} and OpenStreetMap \footnote{\href{https://github.com/osm-search/Nominatim}{https://github.com/osm-search/Nominatim}} to know the size of a location and whether a candidate is contained within another geographic candidate. The size of a location is normalized logarithmically, using a minimum area of 225 m² (the size of a small property) and a maximum area of 530,000 km² (a size covering most countries). 
A candidate that is contained in another and has a small area in square meters will receive a higher score than a candidate that encompasses others and is very large. 

\subsubsection{``Why'' Candidates}
\begin{table}[H]
\centering
\caption{Weights and scoring factors for ``why'' candidates.}
\renewcommand{\arraystretch}{1.4} 
\setlength{\tabcolsep}{10pt} 
\begin{tabular}{>{\centering\arraybackslash}m{3cm}|m{1.5cm} | m{4cm}}
\hline
\textbf{i} & \textbf{$W_qi$} & \textbf{$S_qi$} \\
\hline
Major Causal Entities & 0.35 & $\frac{\text{major\_conj\_count}(c)}{\max_{c' \in C}(\text{major\_conj\_count}(c')}$ \\
Minor Causal Entities & 0.25 & $\frac{\text{minor\_conj\_count}(c)}{\max_{c' \in C}(\text{minor\_conj\_count}(c')}$ \\
Position & 0.20 & $1 - \frac{Position(c)}{N_\text{sentence}-1}$ \\
Causal Verbs & 0.15 & $\frac{\text{causal\_verb\_count}(c)}{\max_{c' \in C}(\text{causal\_verb\_count}(c')}$ \\
Q\&A Similarity & 0.05 & $0$ or $1$ \\
\hline
\end{tabular}
\label{tab:weights_scoring}
\end{table}
The scoring of ``Why'' candidates is primarily based on the detection of causal verbs, conjunctions, and conjunctive phrases that serve as markers of explanatory sentences in French. We distinguish between three types of markers. The most important are major causal entities, which are conjunctions and conjunctive phrases that unequivocally indicate causation, such as \textit{car} (for), \textit{en raison de} (due to), or \textit{parce que} (because). 
By contrast, minor causal entities are expressions that can indicate causation but can also have other meanings, such as \textit{grâce à} (thanks to) or \textit{sous l’effet de} (under the effect of). 
And third, we have causal verbs, such as \textit{induire} (to induce) or \textit{causer} (to cause).
We compute a normalized score for each type of marker as a ratio of the candidate with the highest number of markers. 

\subsubsection{``How'' Candidates}
\begin{table}[H]
\centering
\caption{Weights and scoring factors for ``how'' candidates.}
\renewcommand{\arraystretch}{1.4} 
\setlength{\tabcolsep}{10pt} 
\begin{tabular}{>{\centering\arraybackslash}m{3cm}|m{1.5cm} | m{4cm}}
\hline
\textbf{i} & \textbf{$W_qi$} & \textbf{$S_qi$} \\
\hline
Verb Tense & 0.45 & future $= 0.33$; gerond$= 0.66$; both $= 1.00$; none $=0.00$\\
Copulative Phrases & 0.30 & $\frac{\text{copulative\_phrase\_count}(c)}{\max_{c' \in C}(\text{copulative\_phrase\_count}(c'))}$ \\
Prepositions & 0.20 & $0$ or $1$ \\

Q\&A Similarity & 0.05 & $0$ or $1$ \\
\hline
\end{tabular}
\label{tab:weights_scoring}
\end{table}

Much like with the ``why'' candidates, the ``how'' candidates are evaluated based on the occurrence of certain marker words. 
We look for copulative phrases that convey explanations, such as \textit{au moyen de} (by means of) or \textit{grâce à} (thanks to). 
We also look for the gerund form, which is often used to express a means or method. It is characterized by a verb in the present participle preceded by a preposition, such as \textit{en appelant} (by calling) or \textit{en construisant} (by building). Alternatively, an article can describe a current problem requiring resolution in the future, which is indicated by the future tense. 
We thus count both present participle and future tense verbs, but count prepositions separately to give a preference to the gerund form.

\subsection{Candidate Selection Phase}

Once the candidates are scored and ranked, we must select the candidates to be returned as answers. Selecting only the top candidate is insufficient, as multiple responses can be relevant to a given question. 
Determining how many candidates the system should return for each question is a challenge, as we found even human annotators do not always agree on this matter. 
The solution we opted for is to set a minimum score threshold, and return all candidates above this threshold. This gives us a level of control over the output: the higher this threshold, the fewer candidates will be returned. We tested multiple thresholds and picked the optimal one per question as the one that allows the algorithm to select the set of answer that best matches that returned by humans. This process is detailed in Section \ref{sec:numresults}

\section{Québec News Dataset and Annotation}
\label{dataset}

To conduct our research, we developed the first annotated 5W1H dataset of Québec French-language news articles. We collected 250 
articles from four major media outlets: \textit{La Presse}, \textit{Le Devoir}, \textit{Le Journal de Québec}, and \textit{Radio-Canada}. 
We selected exclusively informational articles, which are the primary target of the 5W1H challenge, and excluded non-informational content such as editorials, reviews, and letters. Only the title and body text of each article were retained, while all other elements, such as metadata, images, and captions, were discarded. 

A key challenge in annotating this dataset lies in the subjective nature of the 5W1H questions. 
Indeed, two individuals may interpret the same article differently and provide different responses to a question. 
Consequently, we hired four annotators who independently read the articles in our dataset and identified the content they deemed to best answer each of the 5W1H questions. These annotators were selected from the student population of our institution. They were given a clear instruction guide and two annotated articles as examples. 
They were instructed to copy the minimal amount of content from the article that contains the answer, without editing or altering it. The annotators were allowed to provide multiple answers to each question, or none if no answer was found in the article. As a result, the answer to each question provided by each annotator consists of a list of passages directly extracted from the article that they feel addresses the target question. The set of annotated articles is available on our GitHub project\footnote{Link removed for blind peer-review.}.

\subsection{Agreement Between Annotators}
To check how much variety there is between the answers selected by different annotators to a given question in a given article, we compute inter-annotator agreement as the ratio of similar answers between them (using our previous measure for similar candidates) to the total number of answers provided by both of them. For instance, if one annotator answered {Justin Trudeau} 
and another answered {François Legault, Justin Trudeau} to the ``who'' question of an article, the similarity score would be $\frac{|\{Justin\ Trudeau,\ Justin\ Trudeau\}|}{|\{Justin\ Trudeau,\ François\ Legault,\ Justin\ Trudeau\}|} = \frac{2}{3} = 66.66\%$, corresponding to two similar answers out of a total of three.  

\begin{center}
\begin{figure}[ht]
    \centering
    \includegraphics[scale=0.5]{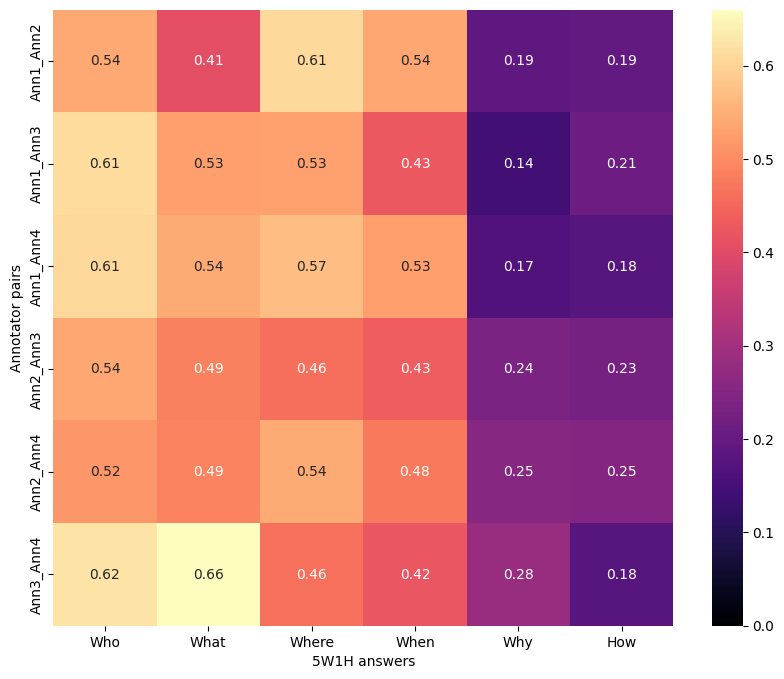}
    \caption{Agreement between annotators per question.}
    \label{fig:anno_anno}
\end{figure}
\end{center}

Using this formula, we can compute the agreement between pairs of annotator for each article, then average over all articles to get their agreement for each 5W1H question. The results for each question and each pair of annotators is given in Figure \ref{fig:anno_anno}.

These results confirm the subjectivity of the 5W1H task: the average agreement between annotators is 0.42, meaning that humans agree on less than half the answers. 
Even the best result, the agreement of annotators 3 and 4 on question ``what'', is only of 0.66, meaning one-third of their answers are different. 

The results also show a clear distinction between ``easy'' and ``hard'' questions to answer. All annotator pairs achieve an average agreement between 0.41 and 0.66 for the ``who'', ``what'', ``where'' and ``when'' questions. These questions ask for a clear piece of information that can be easily isolated in the news article. Most people, it seems, can roughly agree on ``who did what where and when'' in the story. On the other hand, the agreement between annotators is much lower for the ``why'' and ``how'' questions, ranging between 0.14 and 0.28. This shows that determining the cause or manner of an event can be difficult for humans, a problem also observed in \cite{hamborg2019giveme5w1h}. Moreover, these elements may be unclear, taken for granted, or implicitly described in the article \cite{das20125w}, adding to the challenge.




\section{Results and Analysis}
\label{results}
\subsection{Baseline}
As a baseline to compare our algorithm to, we opted for the only other option available in French, using a generative AI to extract answers to the 5W1H questions. 
For this purpose, we chose the GPT-4o model by OpenAI. 
The prompt implements a one-shot learning approach, by giving the system a sample article with the expected answers. We used one of the two sample articles given to the annotators. The prompt used is detailed in the Appendix.

\subsection{Agreement}
\begin{center}
\begin{figure}[ht]
    \centering
    \includegraphics[scale=0.35]{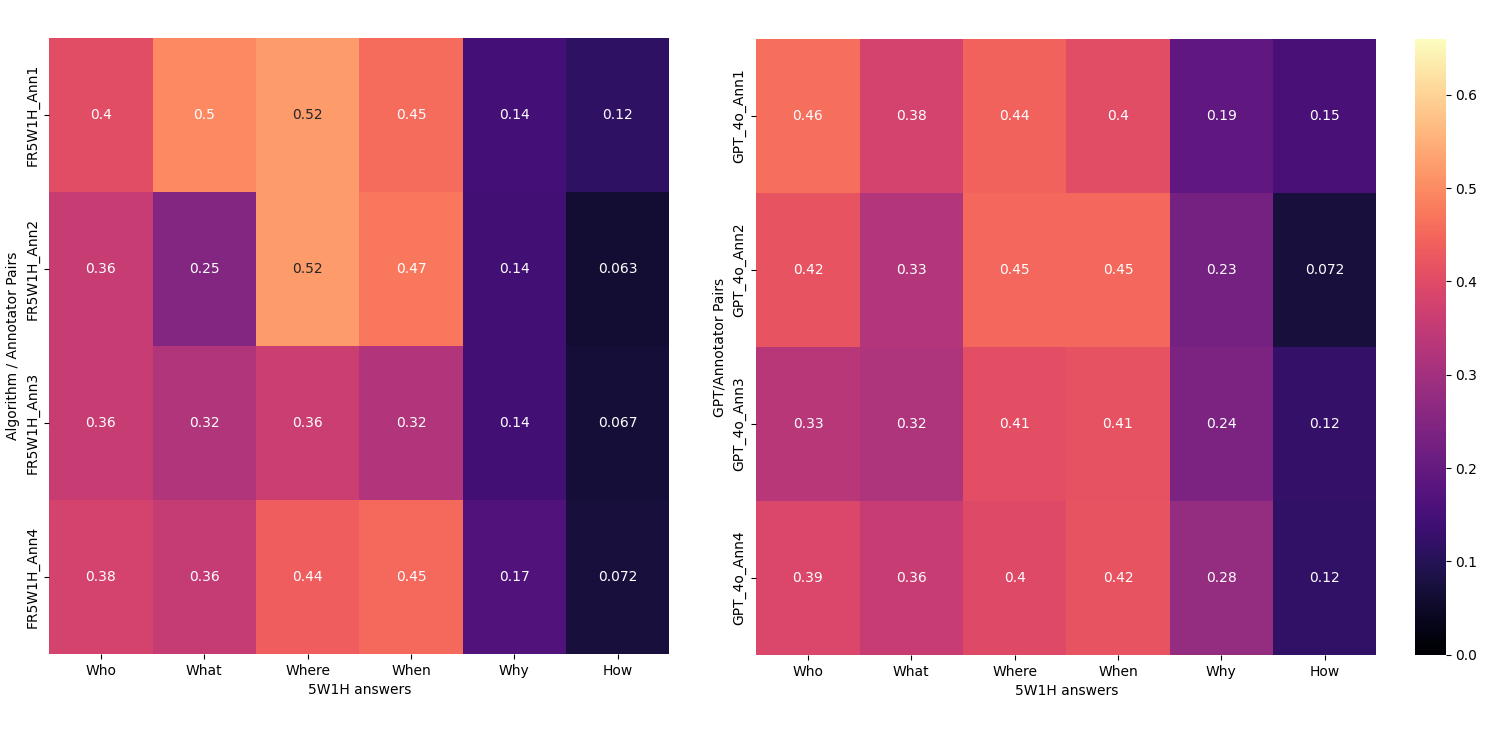}
    \caption{Agreement between our algorithm and each annotator (left) and between GPT-4o and each annotator (right).}
    \label{fig:algo_anno}
\end{figure}
\end{center}

Since our 5W1H annotations showed that there are no gold-standard answers to these questions, we cannot simply evaluate our systems using classical metrics like answer precision or recall. 
Instead, we compute the agreement between each system and the annotators. The idea is to see to what extent these systems perform in a manner equivalent to a human annotator. We used the same technique as to compute the agreement between annotators. The results are illustrated in Figure \ref{fig:algo_anno}. 







These results follow the same trend as between the annotators: the easier first four W questions have a higher average agreement than the harder ``why'' and ``how'' questions. 
However, we also see a significant decrease in the average agreement rates for both solutions: the average agreement with the annotators is of 31\% for our algorithm and of 32\% for GPT-4o. This shows that it is challenging for any algorithm to mimic human abilities in such a subjective task.

When we compare our algorithm and the baseline together, we can see that our algorithm agrees more with the annotators than GPT-4o for the easy W questions, but that GPT-4o agrees more with them for the two harder questions. This is an interesting result, as it shows that, for the first four questions, our algorithm can match one of the best generative AIs available. The last two questions have answers that are more implicit or complex and thus harder to extract with simple rules, while GPT-4o benefits from its reasoning capacities. But a more fine-grained set of factors and better calibrated weights may be able to bridge that gap as they did with the first four questions.






\subsection{Number of Answers}
\label{sec:numresults}
The number of responses returned by our algorithm directly depends on the score threshold applied for each question type. The optimal threshold value is the one that allows the algorithm to return the set of answers that most closely matches that humans would return for that question. Thus, we tested different thresholds for each question and computed the average agreement at each value.
The results are shown in Figure \ref{fig:results_acc}. These results confirm the importance of selecting the right number of responses, especially for the ``what'', ``where'' and ``when'' questions where a high threshold is needed to select a small set of high-quality answers. The optimal thresholds indicated in this figure are also those used to generate the results of Figure \ref{fig:algo_anno}. 

As explained previously, the annotators were instructed to mark all relevant answers for each answer, and our rule-based algorithm returns all answers above the optimal threshold. Our GPT-4o prompt limits the number of ``who'', ``where'' and ``when'' answers to 3 but allows unlimited ``what'', ``why'' and ``how'' answers. It is thus interesting to compare the annotators and algorithms on the number of answers returned. This result is shown in Table \ref{tbl_results}. 

We can see that, even with the freedom to pick any number of answers, the average number of answers provided by human annotators remains between 0.82 and 1.80. In other words, the annotators consistently prefer to give one answer per question. By contrast, the two algorithms return on average 2 and sometimes 3 answers per question, with GPT-4o returning a few more answers on average than our rule-based algorithm.

\begin{center}
\begin{figure}[ht]
    \centering
    \label{fig:results_acc}
    \includegraphics[scale=0.5]{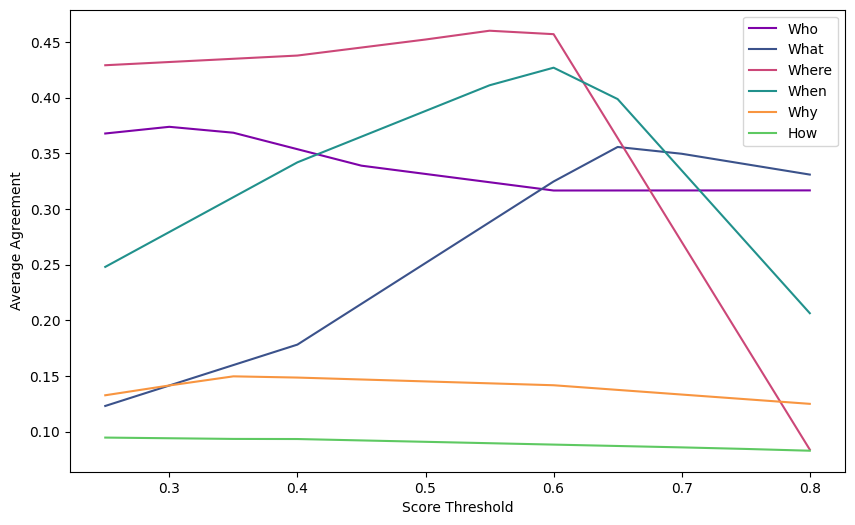}
    \caption{Comparison of average agreement per threshold value for each question type.}
    \label{fig:results_acc}
\end{figure}
\end{center}

\begin{center}
\begin{table}[ht]
\caption{Average number of answers per question}
\begin{tabular}{|c|c c c c c c|}
\hline
Question&Annotator 1& Annotator 2& Annotator 3& Annotator 4& GPT-4o&Algorithm\\
\hline \hline
Who&1.60&1.37&1.17&1.80&3.19&1.62\\
What&1.01&1.75&1.08&1.22&2.84&2.69\\
Where&1.03&1.24&1.20&1.21&2.28&2.83\\
When&0.82&0.98&1.10&0.90&2.24&1.74\\
Why&0.87&0.89&1.23&1.50&2.30&3.01\\
How&0.93&0.85&1.14&1.04&2.69&1.71\\
\hline \hline
All questions&1.05&1.18&1.15&1.28&2.59&2.27\\
\hline
\end{tabular}
\label{tbl_results}
\end{table}
\end{center}

It is interesting to take a closer look at the ``what'', ``where'' and ``why'' results, as they are the three where our algorithm returns the most answers. The first two are questions where our algorithm has a better answer agreement with the annotators than GPT-4o, while the third is one where GPT-4o's answers are clearly better than ours. GPT-4o also returns more answers on average than our algorithm for the ``what'' question, while the opposite is true for the other two questions. This seems to indicate we shouldn't put too much stock in the number of answers, as neither more nor fewer answers seem to be indicative of better agreement with the annotators.

\section{Conclusion}
\label{conclusion}
In this paper, we designed the first ever algorithm to automatically extract answers to 5W1H journalistic questions from French-language news articles. We also made available the first annotated 5W1H corpus of French-language news articles. The algorithm we propose is a weighted-rule-based extraction pipeline. Despite its simplicity, we show that the answers returned by our system agree with those chosen by humans as well as answers selected by GPT-4o, with the added benefits of being completely transparent and having explainable results. We believe that simple improvements, such as better weights and better extraction rules for the ``why'' and ``how'' questions, will allow our algorithm to surpass GPT-4o. Our algorithm could also serve as a tool in future research to analyze the journalistic structure of a news article for various applications, ranging from news aggregators and writing aid tools to fake news detectors.

\section*{Acknowledgements}
This research was made possible by a research grant from the Natural Sciences and Engineering Research Council of Canada (NSERC).



\printbibliography[heading=subbibintoc]

@book{Ross2005,
    author = {Line Ross},
    title = {L’écriture de presse},
    publisher = {Gaëtan Morin Éditeur},
    year = 2005,
    edition=2
}

@book{MartinLagardette2005,
    author = {Jean-Luc Martin-Lagardette},
    title = {Le guide de l'écriture journalistique},
    publisher = {Éditions La Découverte},
    year = 2005,
    edition=6
}

@inproceedings{wang2010chinese,
  title={Chinese news event 5w1h elements extraction using semantic role labeling},
  author={Wang, Wei and Zhao, Dongyan and Wang, Dong},
  booktitle={2010 Third International Symposium on Information Processing},
  pages={484--489},
  year={2010},
  organization={IEEE}
}

@inproceedings{yaman2009classification,
  title={Classification-based strategies for combining multiple 5-w question answering systems},
  author={Yaman, Sibel and Hakkani-T{\"u}r, Dilek and Tur, Gokhan and Grishman, Ralph and Harper, Mary and McKeown, Kathleen R and Meyers, Adam and Sharma, Kartavya},
  booktitle={Tenth Annual Conference of the International Speech Communication Association},
  year={2009}
}

@article{sharma2013news,
  title={News event extraction using 5W1H approach \& its analysis},
  author={Sharma, Smriti and Kumar, Rajesh and Bhadana, Pawan and Gupta, Sumita},
  journal={International Journal of Scientific \& Engineering Research},
  volume={4},
  number={5},
  pages={2064--2068},
  year={2013}
}

@article{parton2009comparing,
  title={Who, what, when, where, why? comparing multiple approaches to the cross-lingual 5w task},
  author={Parton, Kristen and McKeown, Kathleen and Coyne, Robert Eric and Diab, Mona T and Grishman, Ralph and Hakkani-T{\"u}r, Dilek and Harper, Mary and Ji, Heng and Ma, Wei Yun and Meyers, Adam and others},
  year={2009}
}

@article{hamborg2019giveme5w1h,
  title={Giveme5w1h: A universal system for extracting main events from news articles},
  author={Hamborg, Felix and Breitinger, Corinna and Gipp, Bela},
  journal={arXiv preprint arXiv:1909.02766},
  year={2019}
}

@inproceedings{wang2010extracting,
  title={Extracting 5W1H event semantic elements from Chinese online news},
  author={Wang, Wei and Zhao, Dongyan and Zou, Lei and Wang, Dong and Zheng, Weiguo},
  booktitle={International Conference on Web-Age Information Management},
  pages={644--655},
  year={2010},
  organization={Springer}
}

@inproceedings{das20125w,
  title={The 5w structure for sentiment summarization-visualization-tracking},
  author={Das, Amitava and Bandyaopadhyay, Sivaji and Gamb{\"a}ck, Bj{\"o}rn},
  booktitle={International Conference on Intelligent Text Processing and Computational Linguistics},
  pages={540--555},
  year={2012},
  organization={Springer}
}

@book{goffman1974frame,
  title={Frame analysis: An essay on the organization of experience.},
  author={Goffman, Erving},
  year={1974},
  publisher={Harvard University Press}
}

@article{lewis2019mlqa,
  title={MLQA: Evaluating cross-lingual extractive question answering},
  author={Lewis, Patrick and O{\u{g}}uz, Barlas and Rinott, Ruty and Riedel, Sebastian and Schwenk, Holger},
  journal={arXiv preprint arXiv:1910.07475},
  year={2019}
}

@article{chakma2019deep,
  title={Deep semantic role labeling for tweets using 5W1H},
  author={Chakma, Kunal and Das, Amitava and Debbarma, Swapan},
  journal={Computaci{\'o}n y Sistemas},
  volume={23},
  number={3},
  pages={751--763},
  year={2019},
  publisher={Centro de Investigaci{\'o}n en computaci{\'o}n, IPN}
}

@inproceedings{norambuena2020evaluating,
  title={Evaluating the inverted pyramid structure through automatic 5w1h extraction and summarization},
  author={Norambuena, Brian and Horning, Michael and Mitra, Tanushree},
  booktitle={Computational Journalism Symposium},
  year={2020}
}

@article{DBLP:journals/corr/abs-1911-03894,
  author       = {Louis Martin and
                  Benjamin Muller and
                  Pedro Javier Ortiz Su{\'{a}}rez and
                  Yoann Dupont and
                  Laurent Romary and
                  {\'{E}}ric Villemonte de la Clergerie and
                  Djam{\'{e}} Seddah and
                  Beno{\^{\i}}t Sagot},
  title        = {CamemBERT: a Tasty French Language Model},
  journal      = {CoRR},
  volume       = {abs/1911.03894},
  year         = {2019},
  url          = {http://arxiv.org/abs/1911.03894},
  eprinttype    = {arXiv},
  eprint       = {1911.03894},
  bibsource    = {dblp computer science bibliography, https://dblp.org}
}

@book{bleyer1923newspaper,
  title={Newspaper writing and editing},
  author={Bleyer, Willard Grosvenor},
  year={1923},
  publisher={Houghton Mifflin}
}

@article{osti_10274096,
place = {Country unknown/Code not available}, title = {Evaluating the Inverted Pyramid Structure through Automatic 5W1H Extraction and Summarization}, url = {https://par.nsf.gov/biblio/10274096}, abstractNote = {The inverted pyramid is a basic structure of news reporting used by journalists to convey information and it is considered a key element of objectivity in news reporting. In this article, we propose the Inverted Pyramid Scoring method to evaluate how well a news article follows the inverted pyramid structure using main event descriptors (5W1H) extraction and news summarization. We evaluate our proposed method on a proprietary data set of Associated Press news articles across breaking and non-breaking news spanning two topics—political and business. Our results show that the method works at distinguishing the structural differences between breaking and non-breaking news. In particular, our results confirm that breaking news articles are more likely to follow the inverted pyramid structure.}, journal = {Computational Journalism C+J}, author = {Keith, Brian and Horning, Michael and Mitra, Tanushree}, editor = {null}, year = {2020} }

\section*{Appendix}
\subsection*{Q\&A Module Prompts}

\textbf{who}
\begin{itemize}
    \item \texttt{Which person or company is the main subject of this event?}
\end{itemize}

\textbf{what}
\begin{itemize}
    \item \texttt{What is happening to \textit{who\_answer} in this news article? The answer is in the opening sentences.} (if the ``who'' score is $\geq 0.5$)
    \item \texttt{What is the main event? The answer is in the opening sentences.} (otherwise)
\end{itemize}

\textbf{when}
\begin{itemize}
    \item \texttt{When do the events described in the news article take place?}
\end{itemize}

\textbf{where}
\begin{itemize}
    \item \texttt{Where does this happen?}
\end{itemize}

\textbf{why}
\begin{itemize}
    \item \texttt{Why \textit{what\_answer}?} (if the ``what'' score is $\geq 0.2$)
    \item \texttt{Why does \textit{who\_answer} act?} (else if the ``who'' score is $\geq 0.5$)
    \item \texttt{Why did the events detailed in this news article occur?} (otherwise)
\end{itemize}

\textbf{how}
\begin{itemize}
    \item \texttt{How does \textit{who\_answer} do \textit{what\_answer}?} (if the ``what'' score is $\geq 0.2$ and the ``who'' score is $\geq 0.5$)
    \item \texttt{How does \textit{who\_answer} act?} (else if the ``who'' score is $\geq 0.5$)
    \item \texttt{In the following news article, what best answers the question ``how?''} (otherwise)
\end{itemize}

\subsection*{GPT-4o Prompt}
\begin{verbatim}
You are a helpful NLP assistant. 
Your task is to analyze a text from a French press article and extract 
answers to the 5W1H questions:
1. Who? 2. What? 3. Where? 4. When? 5. Why? 6. How?

Here are the instructions for your task:
- Provide direct quotations from the text for each question.
- Do not translate or truncate the quotations.
- You may give multiple answers for each question, but only include the most
relevant ones.
- For the "who," "where," and "when" questions, limit your answers to 2 or 
3 quotations at most.
- If two answers are semantically similar, only use the most relevant one.
- Do not provide any answers other than a JSON-formatted response, as 
shown below.
For instance, consider the following press article:
\end{verbatim}
\textit{Here is the example article provided}
 \begin{verbatim}
     The answer given from this article should be the following:
        {
            "who": ["example who answers"],
            "what": ["example what answers"],
            "where": ["example where answers"],
            "when": ["example when answers"],
            "why": ["example why answers"],
            "how": ["example how answers"]
        }
    Remember to include nothing in your output other than a JSON-formatted
    response.
 \end{verbatim}
 \textit{Here is the article from which we want to extract the answers.}
 \begin{verbatim}
     Remember that your answer must only be a completed version of this 
     json output:
            {
            "who": [],
            "what": [],
            "where": [],
            "when": [],
            "why": [],
            "how": []
        }
 \end{verbatim}
\end{document}